%% file: main.tex
\documentclass{article}
\usepackage{iclr2027_conference,times}
\input{preamble}
\input{metadata}

\iclrfinalcopy
\input{appendix/prompt_setup}
\begin{document}
\maketitle
\lhead{Preprint}
\begin{abstract}
\input{sections/abstract}
\end{abstract}

\input{sections/introduction}
\input{sections/background}
\input{sections/analysis}
\input{sections/method}
\input{sections/experiments}

\input{sections/related_work}

\input{sections/conclusion}

\input{sections/limitations}
\input{sections/statements/ai_use}
\input{sections/statements/ethics}
\input{sections/statements/reproducibility}

\bibliography{references}
\bibliographystyle{iclr2027_conference}

\clearpage
\appendix
\input{appendix/validity_midtraining}

\clearpage
\input{appendix/semantic_novelty}
\clearpage
\input{appendix/implementation}
\input{appendix/baselines}
\input{appendix/controlled_datasets}
\input{appendix/additional_results}
\end{document}

%% file: preamble.tex
\usepackage[T1]{fontenc}
\usepackage[utf8]{inputenc}
\usepackage{amsmath,amssymb}
\usepackage{graphicx}
\usepackage{booktabs}
\usepackage{xcolor}
\usepackage{hyperref}
\usepackage{url}
\graphicspath{{figures/}}
\usepackage{fvextra}
\usepackage{capt-of}
\usepackage{booktabs}
\usepackage{xcolor}
\usepackage{colortbl}
\usepackage{tabularx}
\usepackage{multirow}
\usepackage{subcaption}
\usepackage{bbm}
\usepackage[normalem]{ulem}
\usepackage{wrapfig}
\usepackage{microtype}
\usepackage{fontawesome5}
\usepackage{gradient-text}
\definecolor{ResourceBlue}{HTML}{2458A6}

\newcommand{\eg}{\textit{e.g.}}

\definecolor{ResultHeader}{HTML}{E8E8E8}
\definecolor{ResultGroup}{HTML}{F5F5F5}
\definecolor{ResultAvg}{HTML}{F7F2FF}
\definecolor{ResultRule}{HTML}{919191}
\definecolor{ResultDash}{HTML}{B0B0B0}
\definecolor{ResultModel}{HTML}{666666}

\newcommand{\resultdashline}{%
  \noalign{%
    \vskip 2pt
    \hbox to \linewidth{%
      \color{ResultDash}%
      \leaders\hbox{\rule{3pt}{0.4pt}\hskip 3pt}\hfill
    }%
    \vskip 2pt
  }%
}

\newif\ifshowdraftnotes
\showdraftnotestrue


%% file: metadata.tex
\title{Questioning the Questions: Sustaining Self-Evolution in Reasoning Models}

\author{%
\textbf{Jinyuan Li$^{1}$, Chengsong Huang$^{1}$, Langlin Huang$^{1}$, Donghong Cai$^{1}$, Shiping Gao$^{2}$,}\\
\textbf{Yuyi Yang$^{1}$, Jiaxin Huang$^{1}$\thanks{Corresponding author.}}\\[2pt]
\normalfont $^{1}$ Washington University in St. Louis, 
\normalfont $^{2}$ University of Michigan, Ann Arbor\\[2pt]
{\normalfont\texttt{\{ljinyuan,jiaxinh\}@wustl.edu}}\\[4pt]
{\normalfont\normalsize
  \hypersetup{hidelinks}%
  \href{https://jinyuanli0012.github.io/R-Quest/}{%
    \textcolor[RGB]{45,91,160}{\faGlobe}\enspace
    \gradientRGB{Project Page}{45,91,160}{77,77,151}%
  }%
  \hspace{0.6em}%
  {\color[RGB]{185,185,185}%
    \raisebox{-0.12em}{\rule{0.4pt}{1.05em}}%
  }%
  \hspace{0.6em}%
  \href{https://github.com/JinYuanLi0012/R-Quest}{%
    \textcolor[RGB]{77,77,151}{\faGithub}\enspace
    \gradientRGB{Code}{77,77,151}{109,63,142}%
  }%
}
}
\vspace{-2mm}

%% file: sections/abstract.tex
\vspace{-1mm}
Self-evolving reasoning models learn from their own generated questions, yet repeated self-training can lead to performance collapse. In this paper, we investigate why performance deteriorates over successive rounds and how to sustain self-evolution. 
Our analysis identifies two recurring quality problems in self-generated questions: invalid questions and repeated variants of the same mathematical questions. First, invalid questions become more prevalent across rounds, and answer-consistency filtering further increases their proportion in training data. 
Second, existing question diversity controls based on lexical similarity can miss mathematically equivalent questions expressed in different ways, which leads to question diversity collapse in later training rounds.
Building on these findings, we introduce R-Quest, which uses question validity and novelty feedback to guide self-evolution. We first train the solver to recognize and reject invalid questions, then use its judgments to guide questioner rewards and filter solver training data. To avoid question repetition, we use a frozen base model to compare sampled question pairs and provide novelty feedback.
Empirically, our method consistently achieves the highest average performance on 12 benchmarks in mathematical reasoning, general-domain reasoning, and code generation across two model families. Additionally, R-Quest maintains stable performance gains over ten rounds of self-evolution, peaking in the final round and outperforming R-Zero by 17.32 points.
\vspace{-2mm}


%% file: sections/introduction.tex
\section{Introduction}
\label{sec:introduction}

\begin{wrapfigure}[18]{r}{0.415\textwidth}
    \centering \vspace{-6.7mm}
    \includegraphics[width=\linewidth]
    {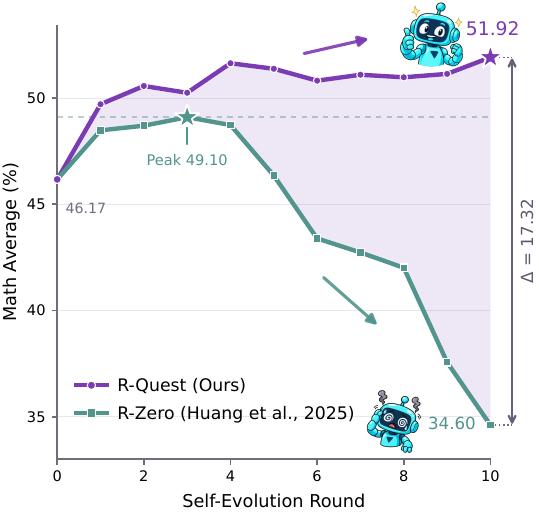} \vspace{-6.5mm}
    \caption{Average performance on seven math benchmarks. R-Quest exceeds the peak of R-Zero in every round and peaks at round ten, leading by \textbf{17.32} points.}
    \label{fig:self_evolution_dynamics1}
\end{wrapfigure}

Self-evolving reasoning models learn by generating and solving their own training tasks~\citep{huang2026g,huang-etal-2023-large,tao2024survey,tan2024large,yu2025guided,sur}, reducing their dependence on human-curated data. Recent frameworks organize this process around questioner and solver roles: the questioner learns to propose challenging problems, while the solver learns to solve them, allowing both roles to co-evolve through reinforcement learning~\citep{zhao2026absolute,chen2026selfquestioning,liu2025spice,shi2026learning}. However, continued self-evolution can reverse early gains and ultimately lead to performance collapse~\citep{huang2025rzero,li2026rdiverse}. In this paper, we study the practical bottlenecks to sustained self-evolution: \emph{Why does performance deteriorate over successive rounds, and how can we address these bottlenecks to achieve sustained self-improvement?}

\begin{figure}[t]
    \centering
    \includegraphics[width=\textwidth]{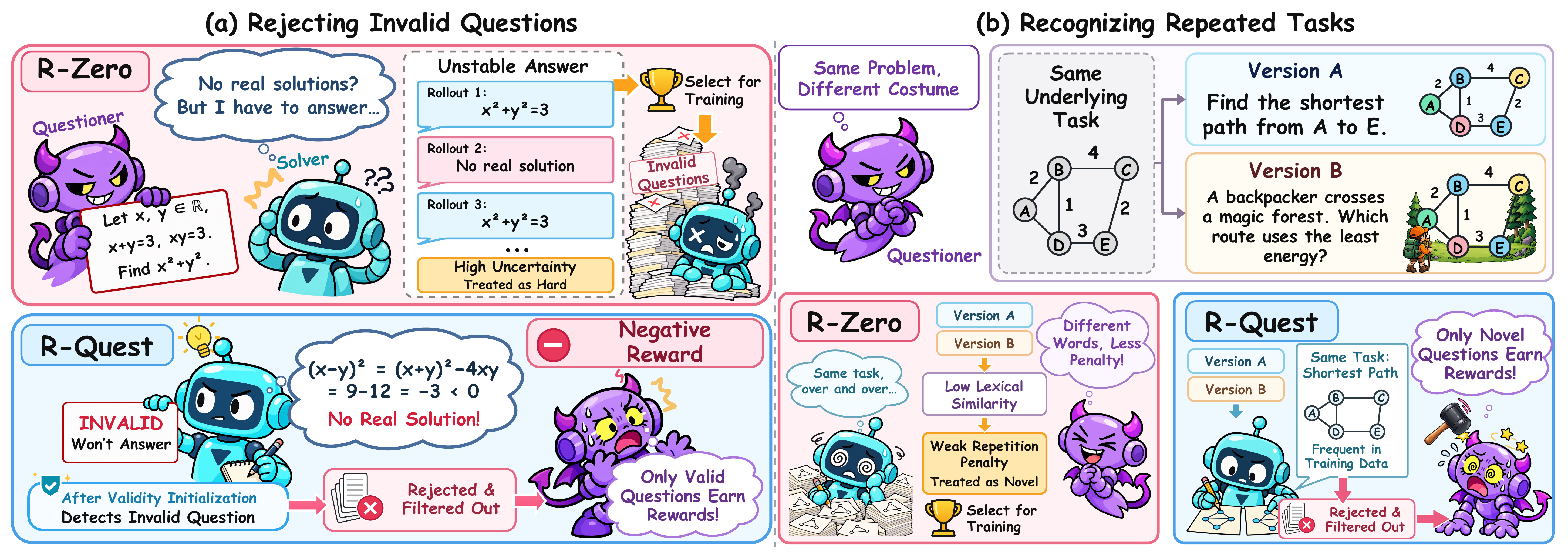}
    \vspace{-6mm}
    \caption{Two question-quality problems in self-evolution and how R-Quest addresses them:
(a) R-Quest can recognize and reject invalid questions;
(b) R-Quest can detect repeated mathematical tasks despite differences
in expression.}
    \label{fig:rquest_overview} \vspace{-5mm}
\end{figure}

To understand why these gains fail to persist, we examine the self-generated questions by R-Zero~\citep{huang2025rzero}.
Our analysis reveals two recurring quality issues: (1) invalid questions and (2) repeated variants of the same mathematical tasks. 
\textbf{First}, some generated questions are invalid (\eg, due to ambiguous definitions, missing conditions or contradictory assumptions), yet the solver is not explicitly trained to reject such questions, as shown in Figure~\ref{fig:rquest_overview}(a).
The loop therefore lacks a mechanism to exclude or suppress invalid questions. 
We find that invalid questions become more prevalent across rounds and that answer-consistency filtering further increases their proportion in training data.
Our comparison study further shows that repairing those invalid questions improves solver performance under matched training settings, indicating that
invalid questions are a contributing factor to performance degradation and
motivating explicit validity feedback during self-evolution.

\textbf{Second}, as shown in Figure~\ref{fig:rquest_overview}(b), the questioner can repeatedly generate variants of the same underlying mathematical task by changing the wording, notation, or numerical parameters, and easily bypass existing repetition penalties. 
We observe substantial concentration in a few mathematical task types in R-Zero's fifth round training data.
Existing diversity controls could split questions with the same task structure into separate clusters. Because repetition penalties depend on cluster size, this fragmentation can weaken the penalty on frequently repeated tasks. These findings suggest that effective diversity control must be grounded in an understanding of what a question asks and how it is solved, so that different expressions of the same task are recognized as repeated training content.

Building on these insights, we introduce R-Quest, which promotes sustained self-evolution by explicitly assessing the validity of generated questions and the novelty of their underlying mathematical tasks. We first train the solver to recognize and reject invalid questions. During self-evolution, we incorporate its validity judgments into the questioner's rewards to discourage invalid generation. 
We also assess task novelty through sampled pairwise comparisons, using the model's understanding of the questions to identify repeated mathematical task structures. Together, these judgments restrict the questioner's difficulty rewards to questions assessed as valid and novel. 
Both feedback signals are generated during subsequent self-evolution without further calls to stronger external models.

Across two model families, R-Quest achieves the highest domain-average scores among evaluated methods on 12 benchmarks spanning mathematics, code generation, and general reasoning. As shown in Figure~\ref{fig:self_evolution_dynamics1}, R-Quest sustains its gains over ten rounds of self-evolution, reaching its best mathematical performance in the final round and outperforming R-Zero by 17.32 points at the same iteration on Qwen3-4B-Base. After ten rounds of self-evolution, R-Quest is 5.75 points above the base model, whereas R-Zero is 11.57 points below it. These results demonstrate that assessing the validity and task novelty of self-generated questions provides a practical route to sustained self-improvement.

%% file: sections/background.tex
\section{Preliminary Study}
\label{sec:preliminary}
\subsection{Background: The Questioner--Solver Framework}
\label{sec:preliminary:background}

Self-evolution for mathematical reasoning commonly adopts a questioner--solver co-training framework. Following the R-Zero paradigm~\citep{huang2025rzero}, both models share the same initialization and are trained alternately using Group Relative Policy Optimization (GRPO)~\citep{shao2024deepseekmath}. For each question $q$ generated by the questioner, the frozen solver samples $M$ responses. Let $\tilde{a}(q)$ be the majority answer and $a_j$ the extracted answer of the $j$-th response; the solver's consistency is defined as $s(q)=\frac{1}{M}\sum_{j=1}^{M}\mathbbm{1}[a_j=\tilde{a}(q)]$. The questioner is trained to generate novel and moderately challenging questions by rewarding uncertainty, with the reward maximized at $s(q)=1/2$, while penalizing repetition. Questions are grouped by lexical similarity using pairwise BLEU distances, and each receives a repetition penalty of $|C(q)|/B$, where $C(q)$ is its cluster and $B$ is the batch size. Subsequently, the frozen questioner proposes new questions, and only those satisfying intermediate consistency ($\vert{}s(q) - 1/2\vert{} \le \delta$) are retained to train the solver. The solver is then updated via GRPO using self-consistency pseudo-labels, assigning binary rewards $\mathbbm{1}[a(y) = \tilde{a}(q)]$ to its responses.

\subsection{Questioning the Questions: Diagnosing Failures in Self-Evolution}
\label{sec:preliminary:diagnosis}

\begin{figure}[t]
\centering \vspace{-5pt}
\captionsetup[subfigure]{skip=0pt}

\begin{subfigure}[t]{0.47\linewidth}
    \centering 
    \includegraphics[width=\linewidth]{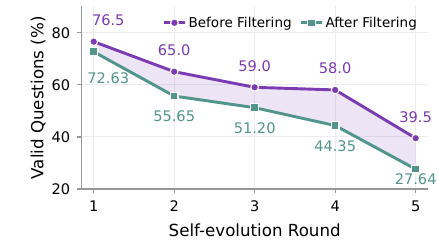}
    \vspace{-6pt}
    \caption{}
    \label{fig:validity_filtering}
\end{subfigure}\hfill%
\begin{subfigure}[t]{0.47\linewidth}
    \centering
    \includegraphics[width=\linewidth]{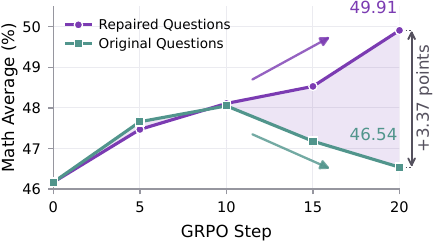}
    \vspace{-6pt}
    \caption{}
    \label{fig:validity_repair_grpo}
\end{subfigure}
\vspace{-5pt}
\caption{Question validity rate and solver performance on Qwen3-4B-Base.
(a) Valid-question rates before and after R-Zero's answer-consistency filter. 
(b) Average math reasoning performance of the solver trained on matched original and repaired datasets.}

\vspace{-14pt}
\label{fig:validity_preliminary}
\end{figure}

\subsubsection{Invalid Questions in Solver Training}
Self-generated questions are not always well posed.
Our investigation into the collapse of R-Zero~\citep{huang2025rzero} reveals a growing proportion of invalid generated questions. We randomly sample 200 questions from the question pool of each of the five R-Zero rounds and use GPT-5.6-Sol to assess their validity. Figure~\ref{fig:validity_preliminary}(a) shows that the valid-question rate falls from 76.5\% in round one to 39.5\% in round five, leaving most questions invalid. 
In addition, R-Zero would filter questions based on majority-voted answer agreement $s(q)$, retaining those between 0.3 and 0.8. We find that among the retained questions, the valid question rate is further reduced, reaching 27.64\% in round five. 
Thus, answer-consistency filtering increases the proportion of invalid questions in solver training data.

\setlength{\intextsep}{2pt} 
\begin{wrapfigure}[19]{r}{0.53\textwidth}
    \centering
    \captionsetup{skip=7pt,belowskip=-8pt}
    \includegraphics[width=\linewidth]{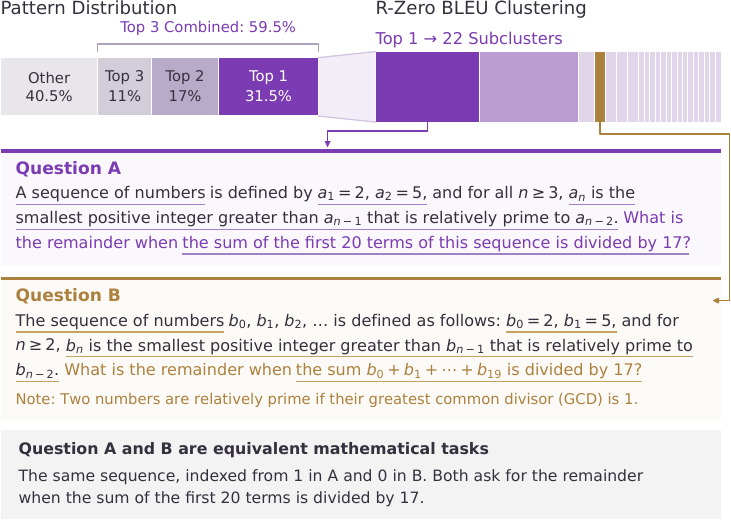}
    \vspace{-3.5mm}
    \caption{Repeated question distribution and BLEU fragmentation in the fifth round of R-Zero. BLEU separates the illustrated mathematically equivalent questions into multiple different clusters.}
    \label{fig:novelty_examples}
\end{wrapfigure}

To test whether invalid questions contribute to solver degradation, we compare GRPO training on matched original and repaired datasets using Qwen3-4B-Base. Each dataset contains 1,273 questions, including the same 766 valid questions. The remaining 507 questions are invalid in the original dataset and replaced by their repaired counterparts in the other. We use GPT-5.6-Sol to repair invalid questions while preserving the domain and minimizing changes in difficulty. Figure~\ref{fig:validity_preliminary}(b) shows comparable early performance followed by diverging trajectories: training on the repaired dataset yields a final average score of 49.91 across seven mathematical benchmarks, compared with 46.54 on the original dataset. These results indicate that invalid training questions are a contributing factor to solver degradation in self-evolution.

\subsubsection{Repetition in Generated Questions}
\label{sec:preliminary:novelty}

Generated questions can repeat the same mathematical task structure despite differences in wording, notation, or numerical parameters. 
Although R-Zero discourages repetition through BLEU-based clustering, lexical similarity may not capture repetition at the level of question types.
In Figure~\ref{fig:novelty_examples}, we show a pair of mathematically equivalent questions in two separate BLEU clusters: they share the same sequence construction and requested computation, despite differences in wording and indexing.

To quantify the prevalence of repeated question types, we use GPT-5.6-Sol to group 200 questions randomly sampled from the filtered question pool generated in R-Zero's fifth round. Questions in each group share the same mathematical task structure, and thus should be considered as ``repeated'' questions. The grouping protocol is detailed in Appendix~\ref{app:exercise-pattern-analysis}. As shown in Figure~\ref{fig:novelty_examples}, the three largest question types account for 59.5\% of the sample. 
Under the BLEU-based clustering procedure of R-Zero, the largest type of question is split into 22 clusters. Such fragmentation can weaken the repetition penalty (which is based on cluster size) on a frequently repeated type. These findings motivate us to assess repetition based on the underlying task structure.

%% file: sections/analysis.tex


%% file: sections/method.tex
\section{Method}

\label{sec:method}

\input{sections/method/overview}
\input{sections/method/validity}
\input{sections/method/novelty}
\input{sections/method/training_loop}

%% file: sections/method/overview.tex
As shown in Figure~\ref{fig:rquest_overview}, R-Quest introduces
two forms of feedback to support sustained self-evolution:
(1) validity feedback from a solver initialized to recognize
and reject invalid questions; and
(2) novelty feedback from the frozen base model, which compares sampled question pairs to identify repeated tasks.
These signals guide questioner rewards during alternating
questioner--solver updates.

%% file: sections/method/validity.tex
\subsection{Validity-Aware Solver Initialization}
\label{sec:method:validity}

Our preliminary analysis in Section~\ref{sec:preliminary:diagnosis} examines invalid questions as a contributor to performance collapse during self-evolution. 
We thus introduce validity-aware solver initialization to teach
the solver to recognize invalid questions before self-evolution.

We construct a dataset by deduplicating questions from the solver training sets of five R-Zero iterations and randomly sampling an equal number from each iteration. We use GPT-5.6-Sol to annotate these questions in two stages. First, the annotator attempts to solve each question and assesses whether it is valid. Second, it independently solves only the valid questions to produce reference answers. We further review these answers and discard examples whose answers cannot be reliably verified. The dataset statistics and annotation protocol are provided in Appendix~\ref{app:validity:data} and \ref{app:validity:annotation}, respectively.

We then train the base model with GRPO to return a mathematical answer for a valid question or \texttt{INVALID} for an invalid one. The model is instructed to attempt a solution and use \texttt{INVALID} only when it identifies a flaw in the question itself. We use the reward
\[
R_\text{init}(q,a)
=
\mathbbm{1}\!\left[a = a^*(q)\right]
-
\lambda\,\mathbbm{1}\!\left[
a = \mathtt{INVALID}
\land a^*(q) \neq \mathtt{INVALID}
\right],
\]
where $a^*(q)$ is the reference answer for a valid question and $\mathtt{INVALID}$ otherwise, and $\lambda > 0$ controls the penalty for rejecting valid questions to discourage excessive refusal. The full reward specification and training settings are provided in Appendix~\ref{app:validity:training}. We initialize the solver with this model, bringing its ability to recognize invalid questions into the self-evolution loop.

%% file: sections/method/novelty.tex
\subsection{Question Novelty Assessment via Sampled Pairwise Judgments}
\label{sec:method:novelty}

As discussed in Section~\ref{sec:preliminary:diagnosis}, question diversity is another dimension along which self-evolution can collapse. We use the frozen base model to perform diversity judgment throughout self-evolution. Two questions are considered the same type when they share essentially the same mathematical setup and ask for the same kind of task. Changes in numerical values or wording alone do not establish a new type, while belonging to the same broad mathematical field is insufficient to establish repetition. The full task specification is provided in Appendix~\ref{app:semantic-novelty:task}.

Let $B$ denote the number of candidate questions in a questioner
rollout batch during GRPO training, and let $p_{ij}\in\{0,1\}$
indicate whether questions $i$ and $j$ are judged to be of the
same type. These binary pairwise judgments define an exhaustive
novelty gate, $\prod_{j\ne i}(1-p_{ij})$, which equals one only
when no repetition is detected: a single same-type match is
sufficient for rejection. However, evaluating this gate for all candidates requires $O(B^2)$ model judgments. 

To reduce this cost, we uniformly sample $K$ other questions per candidate without replacement and reject the candidate if any reference is judged to be of the same type. This reduces the total comparison cost to $O(BK)$ and relaxes
the exhaustive gate, with $K$ controlling how strongly
repeated task types are suppressed. Let $p$ denote the fraction of the remaining questions that the judge would classify as the same type. When $K$ is small relative to $B$, the probability of passing is $\Pr(\mathrm{pass}\mid p)\approx(1-p)^K$. For example, with $K=8$, a same-type fraction of $10\%$ yields a pass probability of approximately $43\%$, compared with only $6\%$ at a fraction of $30\%$. Appendix~\ref{app:semantic-novelty:sampling} provides the exact pass probability and a plot showing how it varies with the
same-type fraction $p$ and the comparison budget $K$.

%% file: sections/method/training_loop.tex
\subsection{Self-Evolution with Validity and Novelty Feedback}
\label{sec:method:loop}

\paragraph{Questioner Rewards.}
We initialize the solver with the validity-aware model from
Section~\ref{sec:method:validity}, bringing its ability to
recognize and reject invalid questions into self-evolution. For a generated question $q$,
it is first instructed  to assess validity while attempting
a solution and return \texttt{INVALID} if it identifies
a flaw in the question. We sample multiple responses and
denote the fraction returning \texttt{INVALID} by $u(q)$.
We define the validity gate as
$g^{\mathrm{valid}}(q)=\frac{1}{2}-u(q)$.
A question is classified as invalid when
$g^{\mathrm{valid}}(q)<0$, meaning that more than half of
the responses return \texttt{INVALID}.

For questions that pass this check, we switch to a standard
problem-solving instruction that asks for
reasoning and a final mathematical answer, and sample a fresh
set of responses. Let $s(q)$ be the proportion of responses
in the largest answer cluster. We define
$g^{\mathrm{novel}}(q)$ to be $1$ if $q$ passes the sampled
novelty check in Section~\ref{sec:method:novelty}, and $0$
otherwise. The questioner reward is
\[
R_{\text{question}}(q)=
\begin{cases}
g^{\mathrm{valid}}(q),
& g^{\mathrm{valid}}(q)<0,\\
g^{\mathrm{novel}}(q)R_{\mathrm{uncertainty}}(q),
& g^{\mathrm{valid}}(q)\geq0,
\end{cases}
\;
R_{\mathrm{uncertainty}}(q)=\min\{s(q),1-s(q)\}.
\]
where $R_{\mathrm{uncertainty}}(q)$ is the uncertainty reward
based on answer consistency. The questioner is updated with GRPO using this reward. 
The prompts for validity assessment and standard problem
solving are provided in the Appendix
Prompts~\ref{app:validity:solver_prompt}
and~\ref{app:self-evolution:math-prompt}, respectively.

\paragraph{Solver Updates.}
The updated questioner generates candidate training questions. We first apply the same majority-vote validity check, then sample mathematical solutions for questions that pass. Consistent with R-Zero~\citep{huang2025rzero}, we construct the training set using answer-consistency filtering and majority-vote pseudo-labels. The solver is updated with GRPO on this dataset, mixed with a small, fixed proportion of the solver initialization data to maintain its ability to recognize and reject invalid questions. Replay examples retain the validity-aware task and reward defined in Appendix~\ref{app:validity:training}. The questioner and solver are then updated alternately. Sampling settings, filtering thresholds, and the replay proportion are provided in Appendix~\ref{app:self-evolution:training}.

%% file: sections/experiments.tex
\section{Experiments}
\label{sec:experiments}
\subsection{Experimental Setup}
\label{sec:experiments:setup}
\paragraph{Models and Baselines.}
We evaluate Qwen3-4B-Base~\citep{yang2025qwen3} and OctoThinker-3B-Hybrid-Base~\citep{wang2025octothinker}, covering Qwen~\citep{yang2025qwen3} and Llama~\citep{grattafiori2024llama} model families. OctoThinker uses mid-training to address the limited gains from applying GRPO directly to Llama base models~\citep{wang2025octothinker}. We compare against the base models, R-Zero, and two recent methods for stabilizing self-evolution, OCNR~\citep{lee2026ocnr} and R-Diverse~\citep{li2026rdiverse}. Baseline descriptions and implementation details are provided in Appendix~\ref{app:baselines}.

\paragraph{Evaluation.}
We evaluate on seven mathematical reasoning benchmarks, three general-domain reasoning benchmarks, and two code generation benchmarks. In line with baselines~\citep{huang2025rzero,lee2026ocnr}, we report mean@32 on AMC and both AIME benchmarks, and greedy-decoding accuracy on the remaining reasoning benchmarks. Code generation is evaluated on HumanEval+~\citep{chen2021evaluating} and MBPP+~\citep{austin2021program} using EvalPlus~\citep{liu2023your}, with pass@1 as the metric. For each domain, we report the checkpoint with the highest average benchmark score across evaluated rounds. Benchmark descriptions and evaluation settings are provided in Appendix~\ref{app:evaluation}.

\paragraph{Training Settings.}
All evaluated methods alternate GRPO updates
of the questioner and solver, using a per-round budget of
10,000 candidate questions, with OCNR requiring twofold
oversampling by design. The solver uses the validity-aware initialization in Section~\ref{sec:method:validity}, with 10\% replay from the initialization dataset each round. The validity gate uses majority voting over nine responses, while the novelty gate uses $K=8$ within-batch comparisons judged by a frozen copy of the corresponding base model. Further training details are provided in Appendices~\ref{app:validity:training} and~\ref{app:self-evolution:training}.

\begin{table}[t]
\centering
\setlength{\belowcaptionskip}{3pt} 
\vspace{-1mm}
\caption{Mathematical reasoning results (\%) across seven benchmarks.
Avg. denotes the mean across benchmarks.
For each backbone, the best score in each column is shown
in \textbf{bold}.}
\vspace{-1.5mm}
\label{tab:main_results}

\footnotesize
\setlength{\tabcolsep}{2.5pt}
\renewcommand{\arraystretch}{1.25}
\setlength{\arrayrulewidth}{0.4pt}
\arrayrulecolor{ResultRule}

\begin{tabularx}{\linewidth}{
    >{\raggedright\arraybackslash}X
    >{\columncolor{ResultAvg}}c
    *{7}{c}
}

\toprule[0.8pt]
\rowcolor{white}
\textbf{Method} &
\textbf{Avg.} &
\textbf{AMC} &
\textbf{Minerva} &
\textbf{MATH} &
\textbf{GSM8K} &
\textbf{Olympiad} &
\textbf{AIME25} &
\textbf{AIME24} \\
\midrule[0.5pt]

\rowcolor{ResultGroup}
\multicolumn{9}{c}{\color{ResultModel}\bfseries\itshape Qwen3-4B-Base} \\
\arrayrulecolor{ResultDash}
\resultdashline
Base model & 46.17 & 47.34 & 51.84 & 74.20 & 88.63 & 39.11 & 11.35 & 10.73 \\
Base model (validity init.) & 47.07 & 49.14 & 53.68 & 74.40 & 89.76 & 41.04 & 10.73 & 10.73 \\
\resultdashline
R-Zero & 49.10 & 54.84 & 54.78 & 76.20 & 92.42 & 42.37 & 12.29 & 10.83 \\
OCNR & 49.57 & 55.94 & 58.09 & \textbf{77.60} & 92.12 & 43.11 & 7.92 & 12.19 \\
R-Diverse & 49.82 & 55.31 & 57.35 & 77.20 & 92.19 & 43.70 & 9.48 & 13.54 \\
\resultdashline
\textbf{R-Quest (our method)} & \textbf{51.64} & \textbf{57.89} & \textbf{58.82} & \textbf{77.60} & \textbf{92.57} & \textbf{46.67} & \textbf{13.44} & \textbf{14.48} \\

\arrayrulecolor{ResultRule}
\midrule[0.5pt]
\rowcolor{ResultGroup}
\multicolumn{9}{c}{\color{ResultModel}\bfseries\itshape OctoThinker-3B} \\
\arrayrulecolor{ResultDash}
\resultdashline
Base model & 24.90 & 25.00 & 23.16 & 41.80 & 69.14 & 11.85 & 0.00 & 3.33 \\
Base model (validity init.) & 25.22 & 19.84 & 26.84 & 43.40 & 69.90 & 13.63 & 0.00 & 2.92 \\
\resultdashline
R-Zero & 28.25 & 25.47 & \textbf{29.78} & 48.80 & 75.89 & 15.11 & 0.10 & 2.60 \\
OCNR
& 28.12 & 30.00 & 28.68 & 49.60 & 73.16 & 15.41 & 0.00 & 0.00 \\
R-Diverse & 28.61 & 29.84 & 27.21 & 49.20 & 74.53 & 16.15 & 0.00 & 3.33  \\
\resultdashline
\textbf{R-Quest (our method)} & \textbf{30.69} & \textbf{32.27} & 29.41 & \textbf{55.80} & \textbf{76.42} & \textbf{17.19} & \textbf{0.31} & \textbf{3.44} \\

\arrayrulecolor{ResultRule}
\bottomrule[0.8pt]
\end{tabularx} \vspace{-3mm}
\end{table}

\subsection{Main Results}
\label{sec:experiments:main}

\begin{wrapfigure}[17]{r}{0.55\textwidth}
\centering
\vspace{-3pt}
\includegraphics[width=\linewidth]
    {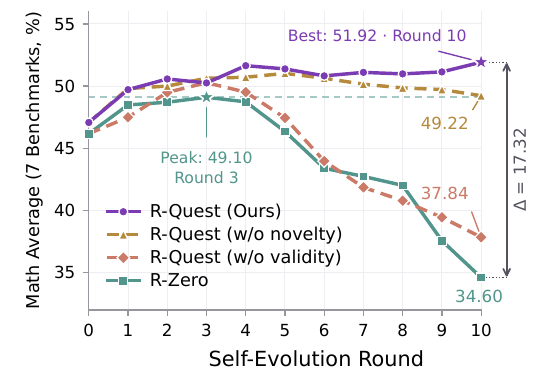}
\captionsetup{skip=-2pt}
\caption{Average performance on seven benchmarks over
ten rounds on Qwen3-4B-Base. Yellow and orange denote R-Quest
without novelty or validity feedback.}
\label{fig:self_evolution_dynamics}
\vspace{-\baselineskip}
\end{wrapfigure}

\paragraph{R-Quest outperforms self-evolution baselines in mathematical reasoning.}
Consistent with evaluations in previous work~\citep{huang2025rzero,li2026rdiverse}, we compare all methods over five rounds of self-evolution, reporting the checkpoint with the highest mathematical average for each method. As shown in Table~\ref{tab:main_results}, R-Quest achieves the highest average performance across seven mathematical reasoning benchmarks on both backbones, surpassing R-Zero, OCNR, and R-Diverse. It improves over Qwen3-4B-Base by 5.47 points and over OctoThinker-3B by 5.79 points. Relative to R-Zero, R-Quest yields additional gains of 2.54
and 2.44 points. Across both backbones, validity-aware initialization yields mathematical performance close to that of the base models.
Most of R-Quest's gains emerge during subsequent self-evolution,
suggesting that initialization supports later learning by
enabling the solver to provide validity feedback.
Measured from the same base models, R-Quest delivers 1.87
and 1.73 times the gains achieved by R-Zero, respectively,
increasing the benefit of self-evolution across Qwen and
Llama model families.

\paragraph{R-Quest sustains gains over extended self-evolution.}
We extend self-evolution to ten rounds to assess sustained gains, as shown in Figure~\ref{fig:self_evolution_dynamics}. R-Zero peaks at 49.10 in round three, then declines to 34.60 by round ten, falling below the base model's score of 46.17. In contrast, R-Quest retains its gains despite minor fluctuations between rounds: every checkpoint from rounds one through ten exceeds R-Zero's best score. R-Quest ultimately reaches its highest score of 51.92 in round ten, exceeding R-Zero by 17.32 points at the same iteration. We further find that either validity or novelty feedback alone achieves a higher peak score than R-Zero. However, validity feedback is more critical for sustained gains: its removal leads to late-stage collapse, whereas the variant without novelty feedback shows a milder decline and remains above the base model.

\begin{table}[t]
\centering
\setlength{\belowcaptionskip}{3pt}
\vspace{-1mm}
\caption{Code generation (pass@1) and general-domain reasoning
results (\%). Avg. denotes the mean within each
domain. Best scores within each backbone are in \textbf{bold}.}
\label{tab:general_code_results}

\footnotesize
\setlength{\tabcolsep}{2.5pt}
\renewcommand{\arraystretch}{1.2}
\setlength{\arrayrulewidth}{0.4pt}
\arrayrulecolor{ResultRule}

\begin{tabularx}{\linewidth}{
    >{\raggedright\arraybackslash}X
    >{\columncolor{ResultAvg}}c
    cc
    >{\columncolor{ResultAvg}}c
    ccc
}
\toprule[0.8pt]
\multirow{2}{*}[-2.8pt]{\textbf{Method}}
& \multicolumn{3}{c}{\textbf{Code generation}}
& \multicolumn{4}{c}{\textbf{General-domain reasoning}} \\
\cmidrule(lr){2-4}\cmidrule(lr){5-8}
& \textbf{Avg.}
& \textbf{HumanEval+}
& \textbf{MBPP+}
& \textbf{Avg.}
& \textbf{SuperGPQA}
& \textbf{MMLU-Pro}
& \textbf{BBEH} \\
\midrule[0.5pt]

\rowcolor{ResultGroup}
\multicolumn{8}{c}{
    \color{ResultModel}\bfseries\itshape Qwen3-4B-Base
} \\
\arrayrulecolor{ResultDash}
\resultdashline
Base model
& 58.79 & 54.88 & 62.70
& 28.79 & 26.21 & 51.41 & 8.74 \\
Base model (validity init.) 
& 59.63 & 55.49  & 63.76 
& 29.40 & 26.28  & 51.58  & 10.35  \\
\resultdashline
R-Zero
& 59.27 & 56.10 & 62.43
& 31.21 & 28.66 & 54.01 & 10.95 \\
OCNR
& 60.28 & 57.32 & 63.23
& 31.56 & 29.05 & 54.82 & 10.82 \\
R-Diverse
& 60.50 & 56.71 & 64.29
& 31.76 & 28.98 & 55.79 & 10.51 \\
\resultdashline
\textbf{R-Quest (our method)}
& \textbf{62.69}  & \textbf{59.76} &  \textbf{65.61}
& \textbf{32.75} & \textbf{29.40} & \textbf{57.36} & \textbf{11.50} \\

\arrayrulecolor{ResultRule}
\midrule[0.5pt]
\rowcolor{ResultGroup}
\multicolumn{8}{c}{\color{ResultModel}\bfseries\itshape OctoThinker-3B} \\
\arrayrulecolor{ResultDash}
\resultdashline
Base model 
& 13.67 & 14.63 & 12.70
& 8.30 & 10.78 & 12.14 & 1.97 \\
Base model (validity init.) 
& 14.77  &  15.24 &  14.29
& 9.04 & 11.42 & 13.53  &  2.17 \\
\resultdashline
R-Zero
& 14.14 & 15.85 & 12.43
& 14.10 & 13.98 & 22.25 & 6.08 \\
OCNR
& 15.60 & 15.85 & 15.34
& 13.83 & 13.89 & 21.86 & 5.73 \\
R-Diverse
& 14.98 & 16.46 & 13.49
& 14.50 & 14.56 & 22.91 & 6.04   \\
\resultdashline
\textbf{R-Quest (our method)}
& \textbf{18.90} & \textbf{17.68} & \textbf{20.11}
& \textbf{15.50} & \textbf{14.96} & \textbf{25.00} & \textbf{6.55} \\

\arrayrulecolor{ResultRule}
\bottomrule[0.8pt]
\end{tabularx} \vspace{-2mm}
\end{table}

\begin{wraptable}{R}{0.5\textwidth}
\centering
\captionsetup{skip=4pt}
\vspace{-2pt}
\caption{Component ablations on Qwen3-4B-Base.
Scores are averaged within each domain.}
\label{tab:gate_ablation}

\footnotesize
\setlength{\tabcolsep}{3pt}
\renewcommand{\arraystretch}{1.08}
\arrayrulecolor{ResultRule}

\begin{tabular*}{\linewidth}{
    @{\extracolsep{\fill}}lccc@{}
}
\toprule[0.8pt]
\textbf{Variant} &
\textbf{Math} &
\textbf{Code} &
\textbf{General} \\
\midrule[0.5pt]
R-Quest
& \textbf{51.64} & \textbf{62.69} & \textbf{32.75} \\
\arrayrulecolor{ResultDash}
\resultdashline
w/o novelty
& 51.02 & 61.98 & 31.80 \\
w/o validity
& 50.27 & 61.28 & 31.53 \\
w/o both (R-Zero)
& 49.10 & 59.27 & 31.21 \\
\resultdashline
Frozen validity judge
& 49.93 & 61.85 & 32.39 \\
\arrayrulecolor{ResultRule}
\bottomrule[0.8pt]
\end{tabular*}
\end{wraptable}

\paragraph{R-Quest improves code generation and general-domain reasoning.}
Prior work has shown that mathematical reasoning training can benefit performance beyond mathematics~\citep{huang2025rzero}. As shown in Table~\ref{tab:general_code_results}, R-Quest substantially improves code generation across both Qwen and Llama backbones. Specifically, using a mathematics-focused self-evolution curriculum, it improves average code generation scores over the base models by 3.90 points on Qwen3-4B-Base and 5.23 points on OctoThinker-3B, whereas R-Zero, OCNR, and R-Diverse yield comparatively modest gains. R-Quest also achieves the highest average performance across the three general-domain reasoning benchmarks on both backbones. Validity initialization alone yields only modest gains in both domains; most of the improvement occurs during subsequent mathematics-focused self-evolution. 
Reducing training on invalid questions may limit the reinforcement of solutions based on unstated assumptions, encouraging reasoning grounded in the information provided across domains. This is particularly relevant to code generation, where correct implementations must satisfy the constraints in the task.

\subsection{Analysis}
\subsubsection{Component Ablations}
\label{sec:experiments:ablations}

\paragraph{Validity and novelty feedback both improve performance.}
As shown in Table~\ref{tab:gate_ablation}, removing either validity or novelty feedback lowers performance across all three domains.  We also compare with a frozen validity judge setting, with novelty feedback enabled. 
This variant underperforms the full R-Quest across all three domains, supporting that continuous validity-training through self-evolution is essential.

\subsubsection{Validity Assessment During Self-Evolution}
\label{sec:experiments:validity}
\begin{figure}[t]
    \centering
    \includegraphics[width=\linewidth]{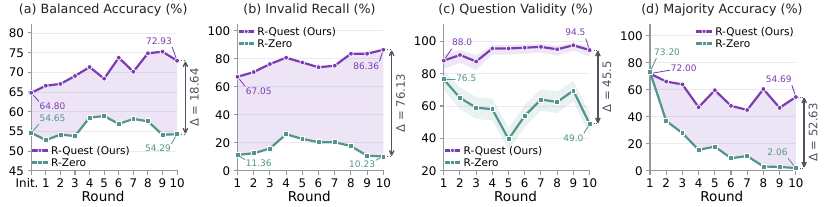}
    \vspace{-8pt}
    \caption{Validity assessment, question and answer quality during
self-evolution on Qwen3-4B-Base.
Panels (a) and (b) show R-Quest's improved validity assessment on held-out
questions. Panels (c) and (d) show that R-Quest maintains high
question validity and better preserves
majority-answer accuracy on valid questions. 
Bands in (c) show pointwise 95\% Wilson confidence intervals.}
    \label{fig:validity_analysis}
\end{figure}

\paragraph{Self-evolution strengthens validity assessment in R-Quest.}
We evaluate each round's solver validity assessment on the same  297 questions held out from the training process, as described in Appendix~\ref{app:validity:data}. Using Prompt~\ref{app:validity:solver_prompt}, each solver is instructed to answer valid questions and reject invalid ones, with one greedy response generated per question. 
We report balanced accuracy as $\frac{1}{2}(\mathrm{Recall}_{\mathrm{valid}}+\mathrm{Recall}_{\mathrm{invalid}})$, where the two terms are the fractions of valid questions accepted and invalid questions rejected, respectively. 
R-Quest's balanced accuracy increases from 64.80\% after initialization to 72.93\% in round ten (Figure~\ref{fig:validity_analysis}(a)).
Since a solver that accepts every question achieves 50\% balanced accuracy without detecting any invalid questions, we additionally report invalid recall to assess this capability directly.
Figure~\ref{fig:validity_analysis}(b) shows that R-Quest achieves an invalid recall of 86.36\% in round ten, whereas recall for R-Zero remains between 10.23\% and 26.14\% across all ten rounds. Thus, the validity assessment capability of R-Quest continues to improve through self-evolution.

\paragraph{R-Quest generates more valid questions.}
We next examine whether stronger solver validity assessment is accompanied by improved question quality. 
For each method and round, we randomly sample 200 questions from the unfiltered pool generated by their questioner. 
We query GPT-5.6-Sol to assess each question’s validity and compute the validity rate. 
Figure~\ref{fig:validity_analysis}(c) shows that R-Quest increases this rate from 88.0\% in round one to 94.5--97.5\% throughout rounds four to ten. In contrast, the rate for R-Zero falls from 76.5\% to 49.0\%. 
This indicates that R-Quest steers question generation toward valid problems as self-evolution proceeds.

\paragraph{R-Quest solver derives more reliable answers.}

Question generation and answer supervision are coupled across
rounds: generated questions shape solver training, and the updated
solver supplies pseudo-labels for subsequent rounds. Among the 200 questions sampled for each method and round, we
evaluate majority-answer accuracy only on those judged valid,
since invalid questions do not provide well-defined mathematical
targets for solver training.
We query GPT-5.6-Sol independently to derive and verify reference answers, and compare the majority-voted answers with these references. 
We report strict accuracy over questions with verified references,
giving no credit to missing answers or unresolved judgments.
Figure~\ref{fig:validity_analysis}(d) shows that the two methods begin with comparable accuracy, 
but by round ten,
R-Quest retains 54.69\% accuracy, whereas R-Zero falls to just 2.06\%. 
This shows that the deterioration in R-Zero extends from question generation to answer supervision. 
By maintaining a predominantly valid question distribution and more reliable pseudo-labels, R-Quest preserves the quality of self-generated supervision that supports sustainable self-evolution.

\subsubsection{Question Diversity During Self-Evolution}
\label{sec:experiments:novelty}

\begin{wrapfigure}{r}{0.60\textwidth}
    \centering
    \includegraphics[width=\linewidth]{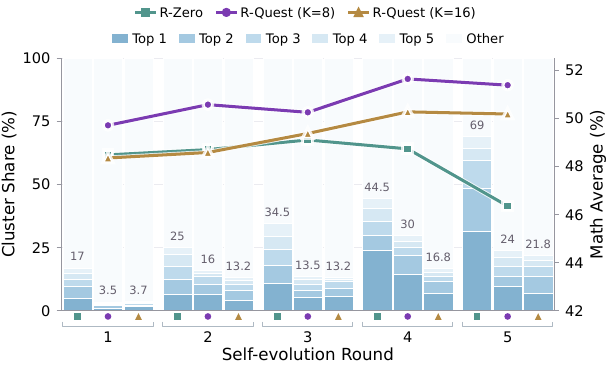}
    \captionsetup{skip=4pt}
    \caption{Question type clusters (bars, left axis) and mathematical performance (lines, right axis) on Qwen3-4B-Base. Numbers above the bars denote top-five group proportion.}
    \label{fig:diversity_dynamics}
    \vspace{-1pt}
\end{wrapfigure}

\paragraph{R-Quest limits the concentration of recurring exercise patterns.}
Following the analysis in Section~\ref{sec:preliminary:novelty}, we uniformly sample 200 solver training question records without replacement for each method and round over the first five rounds. GPT-5.6-Sol sequentially groups them by their underlying mathematical construction and requested task. We measure concentration using the top-five share, the fraction of questions in the five largest groups, averaged over three random orderings of the same sample. Appendix~\ref{app:exercise-pattern-analysis} provides the grouping procedure and task specification. As shown in Figure~\ref{fig:diversity_dynamics}, the top-five share for R-Zero rises from 17.0\% to 69.0\% between rounds one and five. R-Quest with the default $K=8$ remains less concentrated throughout, reaching 24.0\% in round five while maintaining higher performance.

\paragraph{Overly strict diversity control can limit mathematical gains.}
We further examine the strength of novelty control in Figure~\ref{fig:diversity_dynamics} by increasing $K$ from its default value of 8 to 16, with other settings unchanged. A larger $K$ increases the probability that questions from recurring exercise patterns are rejected by the novelty gate (Appendix Figure~\ref{fig:novelty-pass-frequency}). This further lowers the top-five share, most notably from 30.0\% to 16.8\% in round four. However, performance remains below that of $K=8$ throughout, suggesting that overly strict diversity control may suppress more challenging variants within recurring exercise patterns. Effective novelty control should therefore balance exploration of new exercise patterns with progress toward harder problems within existing ones.

%% file: sections/related_work.tex
\section{Related Work}
\label{sec:related_work}

\paragraph{Self-improving Reasoning Models.}
Label-free reinforcement learning uses model confidence,
answer agreement, and entropy as training
signals~\citep{zhao2026learning,zuo2026ttrl,agarwal2026unreasonable}.
Complementary work learns to generate training tasks through
self-play. Absolute Zero validates tasks and solutions through code
execution~\citep{zhao2026absolute}, while R-Zero~\citep{huang2025rzero} and Self-Questioning Language Models~\citep{chen2026selfquestioning} train question generators and solvers using model-derived feedback. Related frameworks ground learning in games, document corpora, vision tasks, and tool environments~\citep{liu2026spiral,liu2025spice,yue2026dr,huang2026skill,huang2026envharness,li2026mm,he2025visplay,dai2026takes,sun2026rethinking}. Our work examines the conditions under which self-generated
questions support sustained improvement, focusing on question
validity and task novelty.

\paragraph{Question Validity and Diversity in Self-evolution.}
For validity, prior work studies question validity in synthetic mathematical data~\citep{liang2026mathdebuggerdetectingdiagnosingerrors,shen2026let,yang2026ttcs}.
For self-evolution, VHG~\citep{lai2026verifier} relies on a stronger external model (GPT-5.4) to verify general mathematical problems during self-play.
We teach the solver to reject invalid questions through mid-training, avoiding calls to stronger external models during subsequent self-evolution.
For diversity, recent methods assess novelty using distances between solver representations or code-embedding similarity~\citep{lee2026ocnr,li2026rdiverse}.
R-Diverse targets repetition in underlying reasoning procedures, but relies on an additional code generator (Qwen2.5-Coder-7B~\citep{hui2024qwen2}) and code encoder (Jina-Code-Embeddings-1.5B~\citep{kryvosheieva2025efficient}) during self-evolution.
This departs from the goal of self-contained self-evolution, where the model generates its own training feedback.
We instead reduce novelty assessment to an atomic judgment of whether two questions are variations of the same mathematical task, allowing a frozen copy of the same base model to provide novelty feedback with only a few random within-batch comparisons.

%% file: sections/conclusion.tex
\section{Conclusion}
\label{sec:conclusion}
We identified invalid questions and repeated mathematical tasks
as two bottlenecks to sustained self-evolution.
R-Quest addresses these bottlenecks by enabling the solver to
recognize and reject invalid questions and using validity and
task-level novelty feedback to guide question generation.
Across two model families and 12 benchmarks, R-Quest achieves
the highest domain-average scores among evaluated methods in
mathematics, code generation, and general reasoning.
It further sustains gains over ten rounds on Qwen3-4B-Base,
reaching its best mathematical performance in the final round.
Further analyses show that R-Quest strengthens validity
assessment, limits task repetition, and maintains more reliable
majority-vote supervision across rounds.
These findings highlight the ability to assess self-generated
questions as a key component of sustained self-improvement.

\subsubsection*{Acknowledgments}

This material is based upon work supported in part by the
National Science Foundation under Award No. 2541822 and
AI for Health Grants. Any opinions, findings and conclusions
or recommendations expressed in this material are those of
the author(s) and do not necessarily reflect the views of
the funding agencies.

%% file: sections/limitations.tex

%% file: sections/statements/ai_use.tex
\subsection*{AI Use Statement}

Generative AI was used in the research workflow for synthetic data generation, model-based annotation and verification, and evaluation, and is also integral to the proposed self-evolution method as described in the paper. It was additionally used for limited language editing. AI-assisted research outputs were reviewed using the validation and consistency-checking procedures described in the paper. The authors reviewed the final manuscript and take responsibility for its contents.

%% file: sections/statements/ethics.tex

%% file: sections/statements/reproducibility.tex
\subsection*{Reproducibility Statement}
We provide the information required to reproduce the main training and evaluation pipeline in the paper and appendices. Appendix~\ref{app:validity} describes construction of the validity-aware initialization dataset, the annotation and reference-answer review procedures, the complete prompts, reward function, and mid-training hyperparameters. Appendix~\ref{app:semantic-novelty} specifies the semantic novelty judgment and sampling procedure. Appendix~\ref{app:self-evolution} provides the question-generation and problem-solving prompts, question-evaluation procedure, solver-data construction, replay strategy, sampling parameters, and GRPO training settings. Appendix~\ref{app:baselines} documents the implementations and configurations of the R-Zero, OCNR, and R-Diverse baselines, and Appendix~\ref{app:evaluation} specifies the evaluation benchmarks, decoding procedures, answer judging, and reported metrics.

%% file: appendix/validity_midtraining.tex
\section{Validity-Aware Mid-Training Details}
\label{app:validity}
\input{appendix/validity/data}
\input{appendix/validity/annotation}

\input{appendix/validity/training}

%% file: appendix/validity/data.tex
\subsection{Dataset Construction}
\label{app:validity:data}

\noindent
\begin{minipage}[t]{0.57\textwidth}
\vspace{0pt}
We collect questions from the solver training sets of the first five R-Zero~\citep{huang2025rzero} rounds using Qwen3-4B-Base~\citep{yang2025qwen3}. After deduplication within and across iterations, we randomly sample 460 questions per iteration, assigning 400 to training and 60 to validation. Annotation and answer review yield a final dataset of 2,240 questions, with statistics reported in Table~\ref{tab:validity:retention}. Table~\ref{tab:validity:invalid_types} reports the distribution of error types among invalid questions.

\end{minipage}\hfill
\begin{minipage}[t]{0.40\textwidth}
\vspace{0pt}
\centering
\setlength{\abovecaptionskip}{0pt}
\setlength{\belowcaptionskip}{5pt}
\captionof{table}{Dataset statistics for validity-aware solver initialization.}\label{tab:validity:retention}
\small
\setlength{\tabcolsep}{3pt}
\begin{tabular*}{\linewidth}{@{\extracolsep{\fill}}lrrr@{}}
\toprule
Split & Valid & Invalid & Total \\
\midrule
Training & 802 & 1,141 & 1,943 \\
Validation & 121 & 176 & 297 \\
\midrule
Total & 923 & 1,317 & 2,240 \\
\bottomrule
\end{tabular*}
\end{minipage}
\par\medskip

\begin{table}[htbp]
\centering
\setlength{\belowcaptionskip}{5pt}
\caption{Error types among the 1,317 invalid questions in the training and validation splits.}
\label{tab:validity:invalid_types}
\small
\begin{tabular}{lrr}
\toprule
Error type & Count & Percentage \\
\midrule
Contradiction       & 380 & 28.9\% \\
No solution         & 362 & 27.5\% \\
Ambiguity           & 315 & 23.9\% \\
Missing information & 181 & 13.7\% \\
Poorly formed questions & 72 & 5.5\% \\
Other               & 7   & 0.5\%  \\
\midrule
Total               & 1,317 & 100\% \\
\bottomrule
\end{tabular}
\end{table}

%% file: appendix/validity/annotation.tex
\subsection{Annotation and Answer Review}
\label{app:validity:annotation}

We use \texttt{gpt-5.6-Sol} to annotate the questions in two stages. Both stages use only the question, without the original R-Zero pseudo-label. First, the model attempts to solve each question and assesses whether it is valid using Prompt~\ref{app:validity:annotation_prompt}. Category A is mapped to \texttt{VALID}, and B--F to \texttt{INVALID}.

\input{appendix/validity/prompt_tex/annotation}

Second, a separate call independently solves the valid questions to produce reference answers using Prompt~\ref{app:validity:answer_prompt}.

\input{appendix/validity/prompt_tex/answer}

We then review the reference answers and exclude questions whose answers cannot be reliably verified. Prompts~\ref{app:validity:review_prompt} and~\ref{app:validity:deep_prompt} provide the review instructions, with the latter used for questions requiring further review. Questions flagged for \texttt{HUMAN\_REVIEW} were manually
reviewed and retained only when their validity and reference
answers could be verified; unresolved cases were excluded.

\input{appendix/validity/prompt_tex/review}
\input{appendix/validity/prompt_tex/deep_review}

%% file: appendix/validity/prompt_tex/annotation.tex
\begin{paperprompt}[label={app:validity:annotation_prompt}]{Validity annotation}
You are a careful reference judge for generated mathematical and reasoning questions. You see only the question. Restate its goal, inspect every definition and condition, check for contradictions and multiple reasonable interpretations, and genuinely attempt a solution. Then classify:

\begin{list}{}{\setlength{\leftmargin}{1.5em}\setlength{\itemsep}{2pt}\setlength{\parsep}{0pt}\setlength{\topsep}{4pt}}
\item[A =] self-contained, valid, solvable, and uniquely gradable;
\item[B =] meaningful but open-ended or not precisely gradable;
\item[C =] missing conditions, multiple answers, or key ambiguity;
\item[D =] contradictory or has no solution;
\item[E =] malformed, undefined, garbled, or unintelligible;
\item[F =] you cannot judge reliably.
\end{list}

Use A only when all strict requirements hold. A question can have multiple solutions and still be A when it explicitly asks for the complete solution set and that set is objectively gradable. Judge only the question as written. Keep the reasoning summary concise but concrete. For A, \texttt{invalid\_type} must be \texttt{null}; for B-F choose the single closest \texttt{invalid\_type}.

\end{paperprompt}

%% file: appendix/validity/prompt_tex/answer.tex
\begin{paperprompt}[label={app:validity:answer_prompt}]{Reference answers}
You are the answer-verification pass for a mathematical reasoning dataset. The validity pass has already classified this question as A (valid). Solve it completely from scratch.

Do not copy or trust an answer hint that may appear in the question. Check every stated condition, substitute or otherwise verify the result, and confirm that the final answer answers exactly what was asked. \texttt{canonical\_final\_answer} must be a concise answer suitable for exact or mathematical-equivalence rewarding, without reasoning or a boxed wrapper.

Set \texttt{answer\_verified}=\texttt{true} only when you have high confidence that the answer is correct and complete. If reliable verification is not possible, set it \texttt{false} and explain the uncertainty; never invent an answer to satisfy the schema.

\end{paperprompt}

%% file: appendix/validity/prompt_tex/review.tex
\begin{paperprompt}[label={app:validity:review_prompt}]{Answer review}
You are an independent mathematical answer consistency auditor. You receive an opaque question \texttt{ID}, the question, a Pass 1 derived answer (which may be missing), and a Pass 2 canonical answer. Neither answer is authoritative.

Read the question carefully and identify exactly what it asks. Independently do the mathematical checks needed to decide whether the canonical answer is correct, complete, and responsive. Decide whether an ambiguity in the question could change the answer. Compare the two supplied answers by mathematical meaning, not merely by string equality: different strings can be equivalent, and equal strings are not proof of correctness. If reliable verification is not possible, say so explicitly. Keep \texttt{reasoning\_summary} concise and concrete; do not provide a long proof.

Relation meanings:

\begin{list}{}{\setlength{\leftmargin}{1.5em}\setlength{\itemsep}{2pt}\setlength{\parsep}{0pt}\setlength{\topsep}{4pt}}
\item[-] \texttt{AGREE}: same answer and meaning.
\item[-] \texttt{EQUIVALENT}: different presentation but mathematically equivalent.
\item[-] \texttt{CONFLICT}: incompatible answers.
\item[-] \texttt{NOT\_COMPARABLE}: their forms or scopes do not permit a reliable direct comparison.
\item[-] \texttt{MISSING}: the derived answer is absent.
\end{list}

Set \texttt{needs\_deep\_review} whenever there is a correctness concern, material ambiguity, conflict, or insufficient basis for a high-confidence screening decision.

\end{paperprompt}

%% file: appendix/validity/prompt_tex/deep_review.tex
\begin{paperprompt}[label={app:validity:deep_prompt}]{Further answer review}
You are the independent deep reviewer for a mathematical answer dataset. You receive only an opaque question \texttt{ID}, the question, a prior derived answer (possibly missing), and a canonical answer (possibly missing). None is authoritative.

Solve the question independently from first principles. Verify what the question actually asks, whether it is clear and valid, and whether your independently derived answer is complete and responsive. Then assess the canonical answer and compare your answer separately with the derived and canonical answers by mathematical meaning, not string equality. Do not assume agreement means correctness. If ambiguity can change the answer, or reliable verification is not possible, say so.

Recommend \texttt{KEEP\_CANONICAL} only when the canonical answer is independently confirmed. Recommend \texttt{REPLACE\_CANONICAL} only when the canonical answer is wrong or missing and your replacement is reliably verified. Recommend \texttt{EXCLUDE} for a materially ambiguous or invalid question. Recommend \texttt{HUMAN\_REVIEW} for unresolved conflicts or uncertainty. Keep \texttt{reasoning\_summary} concise but include the decisive mathematical check. When recommending \texttt{REPLACE\_CANONICAL}, set \texttt{replacement\_answer} to exactly the \texttt{independently\_derived\_answer}; otherwise set \texttt{replacement\_answer} to \texttt{null}. 

\end{paperprompt}

%% file: appendix/validity/training.tex
\subsection{Reward and Training Settings}
\label{app:validity:training}

We train Qwen3-4B-Base~\citep{yang2025qwen3} and OctoThinker-3B-Hybrid-Base~\citep{wang2025octothinker} with
GRPO to return a mathematical answer for a valid question
or \texttt{INVALID} for an invalid one, using
Prompt~\ref{app:validity:solver_prompt}.

\input{appendix/validity/prompt_tex/solver}

Rewards are computed from the final output in \verb|\boxed{}|. Rejection means returning \texttt{INVALID} as the final answer. Mathematical correctness is evaluated against the reviewed reference answer using the \texttt{mathruler} equivalence grader used in R-Zero. For responses with a non-empty final answer box, we apply
the reward in Section~\ref{sec:method:validity}
with $\lambda=0.5$:
\begin{equation}
\label{eq:validity:reward}
r =
\begin{cases}
1, & \text{correct answer or correct rejection},\\
-0.5, & \text{rejection of a valid question},\\
0, & \text{otherwise}.
\end{cases}
\end{equation}
Responses with a missing or empty answer box receive a reward of $-0.1$.

We sample eight responses per question and compute GRPO advantages within each group, with a KL penalty relative to the base model. The checkpoint at step 15 initializes the solver for self-evolution. Table~\ref{tab:validity:training_config} summarizes the training settings.

\begin{table}[t]
\centering
\caption{Training settings for validity-aware solver initialization.}
\label{tab:validity:training_config}
\vspace{6pt}
\begin{tabular}{lr}
\toprule
Setting & Value \\
\midrule
Training steps & 15 \\
Questions per rollout batch & 512 \\
Responses per question & 8 \\
Actor global batch size & 128 \\
Learning rate & $10^{-6}$ \\
KL coefficient & $10^{-2}$ \\
Sampling temperature & 1.0 \\
Top-$p$ & 0.99 \\
Maximum prompt tokens & 2,048 \\
Maximum response tokens & 4,096 \\
\bottomrule
\end{tabular}
\end{table}

%% file: appendix/validity/prompt_tex/solver.tex
\begin{paperprompt}[label={app:validity:solver_prompt}]{Solver training}
Question: \textit{[Question]}

Solve the problem carefully and reason step by step. While solving, check whether the problem itself is well-defined, internally consistent, and solvable as stated.

If the problem is valid, solve it normally. If, while solving, you determine that the problem itself is invalid because of missing information, contradiction, ambiguity, impossibility, malformed conditions, or lack of objective gradability, use \texttt{INVALID} as the final answer.

Do not output \texttt{INVALID} merely because the problem is difficult, because you are uncertain, or because you fail to solve it.

First give your reasoning enclosed within \texttt{\textless{}think\textgreater{}} and \texttt{\textless{}/think\textgreater{}} tags. Then put exactly one final answer inside \texttt{\textbackslash{}boxed\{\}}. For a valid problem, that box must contain the mathematical answer; for an invalid problem, it must contain \texttt{INVALID}.

\end{paperprompt}

%% file: appendix/semantic_novelty.tex
\section{Semantic Novelty Assessment}
\label{app:semantic-novelty}

\subsection{Task Specification}
\label{app:semantic-novelty:task}

The frozen base model compares a candidate question (Question A) with a sampled reference (Question B) using the task specification in Prompt~\ref{app:semantic-novelty:prompt}.

\begin{paperprompt}[label={app:semantic-novelty:prompt}]{Semantic comparison}
You are judging whether two generated math problems are repetitions of the same recurring exercise pattern.

Choose SAME\_TYPE when the two problems have essentially the same distinctive mathematical setup and ask the same kind of task, so that they feel like variations of the same exercise.

Differences in constants, coefficients, variables, formulas, bounds, or other local details do not by themselves make the problems different.

Do not choose SAME\_TYPE merely because the problems share a broad topic, similar wording, presentation style.

If the common pattern is not clear and specific, choose DIFFERENT.

Briefly compare the exercise pattern of the two problems, then end with exactly

\texttt{\textbackslash{}boxed\{SAME\_TYPE\}}

or

\texttt{\textbackslash{}boxed\{DIFFERENT\}}.

Question A:\par
\texttt{\{question\_a\}}

Question B:\par
\texttt{\{question\_b\}}

Analysis:
\end{paperprompt}

\subsection{Effect of Random Sampling}
\label{app:semantic-novelty:sampling}

Consider a candidate among the $B$ successfully parsed questions in a Questioner rollout batch. Each batch initially contains 2,048 generated candidates. Among the other $B-1$ questions, let $m$ be classified as the same exercise type by the judge, with $p=m/(B-1)$. The candidate passes the novelty check only if none of its $K$ references is classified as the same type. Uniform sampling without replacement gives
\begin{equation}
    P_{\mathrm{pass}}(B,m,K)
    = \frac{\binom{B-1-m}{K}}{\binom{B-1}{K}},
    \qquad K \leq B-1,
    \label{eq:novelty-pass-exact}
\end{equation}
where $\binom{n}{K}=0$ for $n<K$. When $K$ is small relative to $B$, the same-type fraction changes little across successive draws, yielding
\begin{equation}
    P_{\mathrm{pass}}(B,m,K) \approx (1-p)^K.
    \label{eq:novelty-pass-approx-appendix}
\end{equation}

\begin{wrapfigure}{r}{0.515\textwidth}
    \vspace{-3pt}
    \centering
    \includegraphics[width=\linewidth]{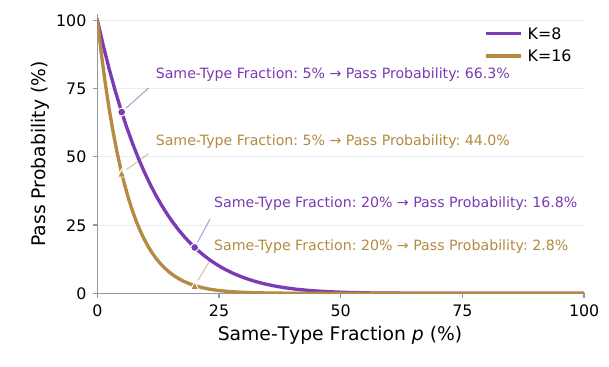}
    \captionsetup{skip=4pt}
    \caption{Approximate pass probability $(1-p)^K$ for the default comparison budget $K=8$ and the additional setting $K=16$.}
    \label{fig:novelty-pass-frequency}
    \vspace{-5pt}
\end{wrapfigure}
Figure~\ref{fig:novelty-pass-frequency} compares the default $K=8$ with the additional $K=16$ setting. Increasing the comparison budget makes same-type matches more likely, raising the probability of rejection. For example, when a recurring exercise pattern accounts for 5\% of the other candidates in the batch, increasing $K$ from 8 to 16 lowers the approximate pass probability from 66.3\% to 44.0\%. For a more frequent pattern accounting for 20\% of the other candidates, the pass probability drops from 16.8\% to just 2.8\%.

Thus, $K=8$ already provides substantial suppression: candidates from a pattern with a 20\% share are rejected about five sixths of the time. Increasing $K$ to 16 raises this rejection probability to 97.2\%, leaving very few candidates of that type eligible for the uncertainty reward. This helps interpret the results in Section~\ref{sec:experiments:novelty}, where $K=16$ lowers the top-five share but yields worse mathematical performance than $K=8$. Overly strict novelty control may suppress not only repeated exercises but also more challenging variants within the same pattern, limiting opportunities for further learning. The default budget therefore balances suppression of recurring patterns with room for continued exploration within them.

\subsection{Exercise-Pattern Concentration Analysis}
\label{app:exercise-pattern-analysis}

We analyze the sampled training records without additional filtering or deduplication. Only question text is supplied to the evaluator; method names, round indices, solver responses, and group sizes are withheld. All methods and rounds use the same evaluator and prompt.

For each ordering, grouping starts with an empty inventory. Questions are processed one at a time using Prompt~\ref{app:exercise-pattern-grouping}. Each existing group is represented by a short pattern description and the question that established it. A question either joins an existing group or establishes a new group with its own description and representative. Existing descriptions and representatives remain fixed throughout the pass. Each question is assigned to exactly one group.

We repeat this procedure from an empty inventory under three random orderings of the same 200 records. For each pass, we sum the sizes of the five largest groups and divide by 200, or sum all groups if fewer than five are formed. We report the mean top-five share across the three orderings.

\begin{paperprompt}[label={app:exercise-pattern-grouping}]{Incremental exercise-pattern grouping}
You are grouping mathematical questions by recurring exercise pattern. Given a new question and an inventory of existing groups, assign it to an existing group or establish a new one.

\medskip
\textbf{Exercise-pattern definition}\par
Questions share a pattern when they have essentially the same distinctive mathematical setup and ask for the same kind of task, making them variations of the same exercise.\par
Differences in wording, notation, numerical values, or other local details do not by themselves establish a new pattern.\par
Changes to defining conditions or the requested task warrant a separate group when they change the underlying exercise.\par
A shared broad topic or generic solution technique is not sufficient.

\medskip
\textbf{Assignment}\par
Compare the new question with the description and representative question of each existing group.\par
Choose the best-matching group only when the shared pattern is clear and specific, without broadening the group's definition.\par
Otherwise, create a new group and briefly describe its mathematical setup and requested task. If the inventory is empty, create a new group.\par
Classify the task as stated rather than repairing it. Apparent flaws alone are not a shared exercise pattern.

\medskip
\textbf{Output}\par
Return a JSON object with the following fields:\par
\texttt{group\_id}: the matching group ID, or \texttt{NEW}.\par
\texttt{pattern}: a concise description for a new group; otherwise, an empty string.\par
\texttt{reason}: a brief explanation of the match or the need for a new group.

\medskip
\textbf{Existing groups}\par
Each entry contains an ID, pattern description, and representative question.\par
\texttt{\{group\_inventory\}}

\medskip
\textbf{New question}\par
\texttt{\{question\}}
\end{paperprompt}

%% file: appendix/implementation.tex
\section{Self-Evolution Settings}
\label{app:implementation}
\label{app:self-evolution}

\subsection{Task Specifications}
\label{app:self-evolution:tasks}

We use the question-generation and mathematical problem-solving instructions from R-Zero~\citep{huang2025rzero}, provided in Prompts~\ref{app:self-evolution:questioner-prompt} and~\ref{app:self-evolution:math-prompt}, respectively. Validity checks use Prompt~\ref{app:validity:solver_prompt}, and semantic comparisons use Prompt~\ref{app:semantic-novelty:prompt}.

\begin{paperprompt}[label={app:self-evolution:questioner-prompt}]{Question generation}
\textbf{System}

You are an expert competition-math problem setter.

FIRST, in your private scratch-pad, think step-by-step to design a brand-new, non-trivial problem. The problem could come from any field of mathematics, including but not limited to algebra, geometry, number theory, combinatorics, prealgebra, probability, statistics, and calculus. Aim for a difficulty such that fewer than 30 \% of advanced high-school students could solve it. Avoid re-using textbook clich\'{e}s or famous contest problems.

THEN, without revealing any of your private thoughts, output **exactly** the following two blocks:

\texttt{\textless{}question\textgreater{}}\par
\texttt{\{The full problem statement on one or more lines\}}\par
\texttt{\textless{}/question\textgreater{}}

\texttt{\textbackslash{}boxed\{final\_answer\}}

Do NOT output anything else---no explanations, no extra markup.

\textbf{User}

Generate one new, challenging reasoning question now. Remember to format the output exactly as instructed.
\end{paperprompt}

\begin{paperprompt}[label={app:self-evolution:math-prompt}]{Mathematical problem solving}
\textbf{System}

Please reason step by step, and put your final answer within \texttt{\textbackslash{}boxed\{\}}.

\textbf{User}

\textit{[Question]}
\end{paperprompt}

\subsection{Sampling and Training Settings}
\label{app:self-evolution:training}

\paragraph{Questioner Rewards.}
To evaluate each generated question, the frozen solver first produces nine validity responses. A question is rejected if more than half of the outputs are \texttt{INVALID}. For questions that pass, we sample ten fresh mathematical solutions to compute answer consistency. These signals, together with the novelty judgment, determine the questioner reward defined in Section~\ref{sec:method:loop}. Questions that fail the required output format receive a reward of $-1$.

\paragraph{Solver Data Construction.}
After updating the questioner, we generate candidate training questions and apply the same evaluation procedure: nine validity responses, followed by ten mathematical solutions for questions that pass. We retain questions with answer consistency $s(q)\in[0.3,0.8]$ and a non-empty majority-vote answer, which serves as the pseudo-label.

\paragraph{Solver Training and Replay.}
Replay examples are sampled from the training split of the mid-training dataset to constitute $10\%$ of the mixed solver dataset. They retain the validity-aware task and reward from Appendix~\ref{app:validity:training}. On newly generated questions, the solver uses the standard mathematical problem-solving task and the R-Zero training reward: $0.9$ for agreement with the pseudo-label under mathematical equivalence checking, plus $0.1$ for placing reasoning inside \texttt{\textless{}think\textgreater{}} and \texttt{\textless{}/think\textgreater{}} tags followed by a boxed answer.

\paragraph{Shared Settings.}
For question evaluation and GRPO rollouts, we use a sampling temperature of 1.0 and a maximum response length of 4,096 tokens. Both the questioner and solver are trained with a learning rate of $10^{-6}$, a KL coefficient of $10^{-2}$, and a maximum prompt length of 2,048 tokens. Table~\ref{tab:self-evolution:sampling} reports question evaluation settings, while Table~\ref{tab:self-evolution:training} reports GRPO training settings. Responses per input in the latter refer to training rollouts.

\begin{table}[htbp]
\centering
\begin{minipage}[t]{0.48\textwidth}
\vspace{0pt}
\centering
\caption{Question evaluation settings.}
\label{tab:self-evolution:sampling}
\vspace{6pt}
\footnotesize
\setlength{\tabcolsep}{3pt}
\begin{tabular*}{\linewidth}{@{\extracolsep{\fill}}lr@{}}
\toprule
Setting & Value \\
\midrule
Validity responses per question & 9 \\
Math solutions per accepted question & 10 \\
Top-$p$ & 1.0 \\
Top-$k$ & 40 \\
Consistency filter for solver data & $[0.3,0.8]$ \\
\bottomrule
\end{tabular*}
\end{minipage}
\hfill
\begin{minipage}[t]{0.48\textwidth}
\vspace{0pt}
\centering
\caption{GRPO training settings.}
\label{tab:self-evolution:training}
\vspace{6pt}
\footnotesize
\setlength{\tabcolsep}{3pt}
\begin{tabular*}{\linewidth}{@{\extracolsep{\fill}}lrr@{}}
\toprule
Setting & Questioner & Solver \\
\midrule
Steps per iteration & 5 & 15 \\
Inputs per rollout batch & 512 & 512 \\
Responses per input & 4 & 5 \\
Actor global batch size & 4 & 128 \\
Top-$p$ & 0.99 & 0.99 \\
\bottomrule
\end{tabular*}
\end{minipage}
\end{table}

%% file: appendix/baselines.tex
\section{Baseline Details}
\label{app:baselines}

\paragraph{Base Models.}
We evaluate Qwen3-4B-Base~\citep{yang2025qwen3} and OctoThinker-3B~\citep{wang2025octothinker} without additional training to establish their performance before self-evolution.

\paragraph{R-Zero.}
We use the released implementation of R-Zero~\citep{huang2025rzero}, which alternates reinforcement learning updates of a questioner and a solver. The questioner reward combines solver uncertainty with a BLEU-based penalty for within-batch repetition. Generated questions are assigned pseudo-labels through majority voting and used for subsequent solver training.

\paragraph{OCNR.}
We implement OCNR~\citep{lee2026ocnr} from its published description within the R-Zero training framework. OCNR augments the questioner reward with a novelty score from a one-class detector trained to distinguish questions used for solver training from questions withheld from training. Following OCNR's twofold overgeneration scheme, we generate 20,000 candidate questions per round, twice our standard budget. The detector maintains one learnable center in the solver representation space. Using values reported in the paper, we set the novelty reward coefficient to $0.3$, the distance temperature $\tau=0.5$, and the EMA update weight $\alpha=0.05$. Question representations are obtained by mean-pooling the solver's final-layer hidden states. The detector receives one full-batch SGD update per round with a learning rate of $7\times10^{-3}$.

\paragraph{R-Diverse.}
We implement R-Diverse~\citep{li2026rdiverse} from its published description within the R-Zero training framework. It penalizes repetition based on how questions are solved, both within the current batch and across rounds. Following the published configuration, we use Qwen2.5-Coder-7B~\citep{hui2024qwen2} with greedy decoding to represent questions as solution programs and Jina-Code-Embeddings-1.5B~\citep{kryvosheieva2025efficient} to measure their similarity within the current batch and against earlier questions. For within-batch repetition, we retain R-Zero's average-linkage clustering with a cosine-distance threshold of $0.5$. Following the paper, we set the coefficients of the within-batch and cross-iteration repetition penalties to $1$. The memory penalty combines maximum and mean cosine similarities with equal weights and thresholds of $0.5$ and $0.25$, respectively. From the second round onward, historical examples are uniformly sampled to constitute $30\%$ of the solver training data.

\clearpage
\section{Evaluation Benchmarks}
\label{app:evaluation}

Although our self-evolution process focuses on mathematical questions, we evaluate both mathematical reasoning and its generalization to other domains. Our evaluation covers mathematical reasoning, general-domain reasoning, and code generation.

\paragraph{Mathematical Reasoning.}
We evaluate on seven benchmarks: AMC, Minerva~\citep{lewkowycz2022solving}, MATH-500~\citep{hendrycks2021measuring}, GSM8K~\citep{cobbe2021training}, OlympiadBench~\citep{he-etal-2024-olympiadbench}, AIME24, and AIME25. We use Qwen3-32B~\citep{yang2025qwen3} to judge the equivalence of model answers and reference answers. Consistent with R-Zero~\citep{huang2025rzero}, we report mean@32 on AMC and both AIME benchmarks, and accuracy under greedy decoding on the remaining benchmarks.

\paragraph{General Domain Reasoning.}
We evaluate generalization beyond mathematics on three benchmarks:
\begin{itemize}
    \item \textbf{MMLU-Pro}~\citep{wang2024mmlu}: Multiple-choice questions across 14 academic domains, requiring both subject knowledge and reasoning.
    \item \textbf{SuperGPQA}~\citep{du2026supergpqa}: Graduate-level multiple-choice questions across 285 disciplines, assessing specialized knowledge and reasoning.
    \item \textbf{BBEH}~\citep{kazemi2025big}: Challenging tasks covering logical deduction, spatial reasoning, and multi-step computation.
\end{itemize}
We follow the R-Zero evaluation protocol and report exact-match accuracy under greedy decoding.

\paragraph{Code Generation.}
We evaluate code generation on two Python programming benchmarks:
\begin{itemize}
    \item \textbf{HumanEval+}~\citep{chen2021evaluating,liu2023your}: Completing Python functions given their signatures and natural-language specifications.
    \item \textbf{MBPP+}~\citep{austin2021program,liu2023your}: Generating short Python programs from natural-language descriptions of basic programming tasks.
\end{itemize}
We use greedy decoding and report pass@1 on the full EvalPlus test suites.

%% file: appendix/controlled_datasets.tex

%% file: appendix/additional_results.tex
\clearpage
\section{Additional Results}
\label{app:additional_results}

\subsection{Detailed Ablation Results}
\label{app:detailed_ablation}


Table~\ref{tab:gate_ablation_full} reports the individual benchmark scores underlying the domain averages in Table~\ref{tab:gate_ablation}.
Removing either validity or novelty feedback lowers performance on both code generation benchmarks and all three general reasoning benchmarks.
The effects vary across mathematical benchmarks, with removing validity feedback producing the largest drop on OlympiadBench.
Although individual ablations achieve higher scores on some benchmarks, the full R-Quest achieves the highest average performance in all three domains.

\begin{table}[t]
\centering
\setlength{\belowcaptionskip}{5pt}
\caption{Per-benchmark results for the gate ablations on Qwen3-4B-Base.}
\label{tab:gate_ablation_full}

\footnotesize
\setlength{\tabcolsep}{2.5pt}
\renewcommand{\arraystretch}{1.18}
\setlength{\arrayrulewidth}{0.4pt}
\arrayrulecolor{ResultRule}

\begin{tabularx}{\linewidth}{
    >{\raggedright\arraybackslash}X
    >{\columncolor{ResultAvg}}c
    *{7}{c}
}
\toprule[0.8pt]
\multirow{2}{*}[-2pt]{\textbf{Variant}}
& \multicolumn{8}{c}{\textbf{Mathematical reasoning}} \\
\cmidrule(lr){2-9}
& \textbf{Avg.}
& \textbf{AMC}
& \textbf{Minerva}
& \textbf{MATH}
& \textbf{GSM8K}
& \textbf{Olympiad}
& \textbf{AIME25}
& \textbf{AIME24} \\
\midrule[0.5pt]

R-Quest
& 51.64 & 57.89 & 58.82 & 77.60 & 92.57 & 46.67 & 13.44 & 14.48 \\
\arrayrulecolor{ResultDash}
\resultdashline
w/o novelty
& 51.02 & 55.94 & 56.62 & 78.00 & 91.81 & 45.63 & 15.42 & 13.75 \\
w/o validity
& 50.27 & 56.48 & 58.82 & 77.00 & 91.74 & 41.19 & 13.54 & 13.12 \\
w/o both (R-Zero)
& 49.10 & 54.84 & 54.78 & 76.20 & 92.42 & 42.37 & 12.29 & 10.83 \\
\arrayrulecolor{ResultDash}
\resultdashline
Frozen validity judge
& 49.93 & 58.13 & 56.99 & 78.20 & 92.12 & 43.11 & 8.02 & 12.92 \\

\arrayrulecolor{ResultRule}
\bottomrule[0.8pt]
\end{tabularx}

\vspace{8pt}

\begin{tabularx}{\linewidth}{
    >{\raggedright\arraybackslash}X
    >{\columncolor{ResultAvg}}c
    cc
    >{\columncolor{ResultAvg}}c
    ccc
}
\toprule[0.8pt]
\multirow{2}{*}[-2pt]{\textbf{Variant}}
& \multicolumn{3}{c}{\textbf{Code generation}}
& \multicolumn{4}{c}{\textbf{General-domain reasoning}} \\
\cmidrule(lr){2-4}\cmidrule(lr){5-8}
& \textbf{Avg.}
& \textbf{HumanEval+}
& \textbf{MBPP+}
& \textbf{Avg.}
& \textbf{SuperGPQA}
& \textbf{MMLU-Pro}
& \textbf{BBEH} \\
\midrule[0.5pt]

R-Quest
& 62.69 & 59.76 & 65.61
& 32.75 & 29.40 & 57.36 & 11.50 \\
\arrayrulecolor{ResultDash}
\resultdashline
w/o novelty
& 61.98  & 59.15  & 64.81 & 31.80 & 29.21 & 55.96 & 10.24\\
w/o validity 
& 61.28 & 58.54 & 64.02 & 31.53 & 28.43 & 55.39 & 10.77\\
w/o both (R-Zero)
& 59.27 & 56.10 & 62.43
& 31.21 & 28.66 & 54.01 & 10.95  \\
\resultdashline
Frozen validity judge
& 61.85 & 59.15 & 64.55 & 32.39 & 29.41 & 56.25 & 11.50 \\

\arrayrulecolor{ResultRule}
\bottomrule[0.8pt]
\end{tabularx}

\end{table}